\documentclass[runningheads]{llncs}
\usepackage{graphicx,subcaption}
\usepackage{amsmath}
\usepackage{amsfonts}
\usepackage{booktabs}
\usepackage{todonotes}
\usepackage{semantic}
\usepackage{multirow}
\usepackage{array}
\usepackage[section]{placeins}
\usepackage{hyperref}
\usepackage[vlined]{algorithm2e}
\usepackage{listings}
\usepackage{amsmath}
\usepackage{amssymb}
\usepackage{tabularx}

\usepackage{color}
\definecolor{keywordcolor}{rgb}{0.7, 0.1, 0.1}   
\definecolor{tacticcolor}{rgb}{0.0, 0.1, 0.6}    
\definecolor{commentcolor}{rgb}{0.4, 0.4, 0.4}   
\definecolor{symbolcolor}{rgb}{0.0, 0.1, 0.6}    
\definecolor{sortcolor}{rgb}{0.1, 0.5, 0.1}      
\definecolor{attributecolor}{rgb}{0.7, 0.1, 0.1} 

\begin{document}
\title{Towards Automated Functional Equation Proving: A Benchmark Dataset and A Domain-Specific In-Context Agent}
\titlerunning{FEAS}

\author{Mahdi Buali\inst{1}\orcidID{0009-0002-2376-8795} \and Robert Hoehndorf\inst{1}\orcidID{0000-0001-8149-5890}}
\authorrunning{Buali \& Hoehndorf}

\institute{Computer Science Program, Computer, Electrical, and
  Mathematical Sciences \& Engineering Division, King Abdullah
  University of Science and Technology, Thuwal 23955, Saudi Arabia\\
\email{\{mahdi.buali, robert.hoehndorf\}@kaust.edu.sa}}

\maketitle              
\begin{abstract}
  Automated Theorem Proving (ATP) faces significant challenges due to
  the vast action space and the computational demands of proof
  generation. Recent advances have utilized Large Language Models
  (LLMs) for action selection in ATP, but these methods often require
  substantial computational resources. This study introduces the
  Functional Equation Automated Solver (FEAS), an agent that builds on
  the COPRA in-context learning framework within the Lean
  environment. FEAS innovates by refining prompt generation and
  response parsing mechanisms, integrating domain-specific heuristics
  for functional equations, and introducing the FunEq dataset—a
  rigorously curated collection of functional equation problems
  categorized into three difficulty levels. The agent's performance is
  evaluated against established baselines using this dataset,
  demonstrating improvements in theorem proving accuracy, particularly
  with the integration of functional equation-specific heuristics. Our
  results highlight the effectiveness of FEAS in generating and
  formalizing high-level proof strategies into Lean proofs,
  emphasizing the potential of tailored approaches in domain-specific
  ATP challenges.
\end{abstract}



\section{Introduction}

Automated theorem proving (ATP) has long been a challenging endeavor
in computer science \cite{ref1}. Formalizing mathematics for efficient
machine processing presents a significant hurdle, further complicated
by the inherent infinite nature of the action space for proof
construction. Interactive theorem provers (ITPs) like Lean
\cite{ref2}, Coq \cite{ref3}, Isabelle \cite{ref4}, and HOL4
\cite{ref5} offer a solution by facilitating formal proofs through
user-guided application of tactics until the desired goals are
achieved.

Recent efforts have explored the use of Large Language Models (LLMs)
as action selectors to address the vast action space in ATP
\cite{ref6} \cite{ref7}. These approaches involve training LLMs from scratch
\cite{ref8} or fine-tuning pre-trained models \cite{ref9} to generate
plausible actions within the context of formal proofs. However, such
methods often incur significant computational costs.

In-context learning, exemplified by the COPRA agent \cite{ref10},
presents a promising avenue to overcome the computational
bottleneck. This approach has demonstrated success in other domains
like machine translation and code generation \cite{ref11}.

Evaluating the capabilities of these algorithms may relies on
challenging problems encountered in high school math Olympiads. The
International Mathematical Olympiad (IMO) \cite{ref12} represents the
pinnacle of difficulty in this domain. Notably, AlphaGeometry
\cite{ref13} recently achieved progress in automated theorem proving
for geometric problems using LLMs with a competitive performance
comparing to IMO participant. However, the field of functional equations, another core topic within IMO's algebraic domain involving finding all unknown functions satisfying specific conditions, remains largely unexplored in the realm of automated theorem proving.

In this project, we build upon the foundation of the COPRA in-context
learning agent \cite{ref10}, working specifically within the Lean
environment, yet, expanding evaluation to various general-purpose
LLMs. Our key contributions include:
\begin{itemize}
\item \textbf{FunEq Dataset}: We created the FunEq
  dataset \footnote{\url{https://github.com/bio-ontology-research-group/FunEq.git}}, a curated
  collection of functional equation problems formalized in Lean. This
  dataset spans three difficulty levels (simple, intermediate, hard),
  providing a rigorous benchmark for evaluating automated theorem
  provers in this domain. Hard problems are drawn from shortlisted IMO
  problems.
\item \textbf{FEAS Agent}: We introduce the FEAS
  agent\footnote{\url{https://github.com/bio-ontology-research-group/FEAS.git}},
  which refines COPRA's prompt generation and response parsing
  mechanisms. FEAS instructs a LLM to produce high-level proof
  strategies in natural language, followed by their formalization and
  translation into Lean proofs. It adopts a robust block-based parsing
  strategy for error handling and backtracking.
\item \textbf{Heuristic Integration}: To enhance and stabilize FEAS's
  performance, we explicitly incorporate domain-specific functional
  equation heuristics \cite{ref14} directly into the agent's prompts.
\end{itemize}

\section{Related Work}

Deep learning has emerged as a promising approach to tackle the
combinatorial explosion of the search space in automated theorem
proving (ATP) \cite{yang2019learning} \cite{blaauwbroek2020tactician} \cite{gauthier2021tactictoe}  \cite{wu2021tacticzero}.  The advent of
Transformer-based language models revolutionized automated theorem
proving by eliminating the need to explicitly hardcode the syntax of
interactive theorem provers (ITPs).  GPT-f \cite{ref6} pioneered this
approach, utilizing language models to generate novel proofs accepted
into the Metamath \cite{ref19} library. PACT \cite{ref7}, a follow-up
project, utilized self-supervised data to improve tactic prediction in
the Lean proof assistant.  Further enhancements with expert iteration
\cite{ref15} enabled autonomous curriculum learning, achieving
state-of-the-art performance on the miniF2F benchmark
\cite{zheng2021minif2f}, a dataset of formal Olympiad-style problems.

Subsequently, Thor \cite{jiang2022thor} integrated language models
with Isabelle's Sledghammers \cite{paulsson2012three} for premise
selection, alleviating the need for explicit specification of every
proof step. Other work \cite{auto} leveraged this integration, employing
in-context learning for autoformalization and expert iteration to
achieve improved results on the MiniF2F benchmark.  Concurrently, HTPS
\cite{ref8} explored the integration of reinforcement learning with
LLMs for guided proof search. 

Recent advances have sought to address the computational cost of LLM
pretraining. LLEMMA \cite{ref9} continued pretraining Code Llama on
mathematical data, demonstrating capability in formal theorem
proving. ReProver \cite{ref16} focused on premise selection using a
retrieval-augmented approach, achieving success with relatively modest
computational resources.

However, the computational burden of fine-tuning LLMs remained a
concern. COPRA \cite{ref10} addressed this by employing
general-purpose LLMs within an in-context learning framework. This
approach repeatedly queries a LLM to propose tactics, leveraging
feedback from the proof environment and retrieved lemmas to refine
subsequent queries.  While COPRA outperformed several baselines, it,
like most prior works, generates proofs one tactic at a time, focusing
on low-level proof steps in comparison to human-like informal
reasoning.  Additionally, previous work primarily developed general
solvers without leveraging domain-specific knowledge, limiting their
efficacy in specialized areas like functional equations.

\section{The FunEq Dataset}

We developed the \textit{FunEq} dataset, a manually curated collection
of functional equation problems formalized in Lean. Our focus on
functional equations is motivated by the fact that, while a
specialized domain, their solutions necessitate a diverse array of
proof techniques. These range from basic algebraic manipulations to
sophisticated reasoning about concepts like continuity~\cite{ref18},
providing a rich testing ground for automated theorem provers.

To accommodate varying levels of difficulty, \textit{FunEq} is
structured into three categories:
\begin{description}
\item[Simple Dataset] This dataset introduces 18 problems which
  require only fundamental functional equation reasoning steps. Proofs
  primarily involve simple substitutions, linear arithmetic, the use
  of involutions, straightforward induction, and basic case analysis.
\item[Intermediate Dataset] This dataset contains 15 problems which
  focuse on proving intermediate lemmas often encountered in the
  solution process of more complex functional equations such as
  establishing injectivity and surjectivity. Problems are sourced
  primarily from Evan Chen's article \cite{ref14} and the book
  ``Functional Equations: A Problem-Solving Approach'' by Venkatachala
  \cite{ref18}.
\item[Hard Dataset] This dataset consists of most of the International
  Mathematical Olympiad (IMO) shortlisted functional equation problems
  since 2002 \cite{ref12}. These problems, originally posed in the
  context of finding all functions satisfying given hypotheses, have
  been reformulated for Lean 3 by explicitly stating the solutions as
  the goal state. This modification simplifies the problem
  representation compared to their original form in the competition.
\end{description}

\section{The FEAS Agent}

\begin{algorithm}[H]
\DontPrintSemicolon
\SetAlgoLined
\SetKwFunction{FMain}{FEAS}
\SetKwProg{Fn}{Function}{:}{}
\Fn{\FMain{$O, \alpha$}}{
    PUSH(\textit{st}, \( O \))\;
    \For{\( j \leftarrow 1 \) \KwTo \( t \)}{
        \If{\( \alpha = \{\} \)}{
            \( p \leftarrow \text{PROMPTIFY}(\textit{st}, \text{Bad}(O)) \)\;
            \( \alpha \leftarrow \text{PARSETACTIC}(\text{LLM}(p)) \)\;
        }
        \( a \leftarrow  \text{POP}(\alpha) \)\;
        \( O' \leftarrow T(O, a) \)\;
        \uIf{\( O' = \text{QED} \)}{
            \textbf{terminate successfully}\;
        }
        \uElseIf{\( O' \in \text{Err} \) \textbf{or} \( \exists O'' \) s.t. \( O' \equiv O'' \)}{
            \text{add} \( a \) \text{to} \text{Bad}(O)\;
            \( \alpha \leftarrow \{\} \)\;
            
        }
        \Else{
            \FMain{\(O', \alpha\)}\;
        }
    }
    POP(\textit{st})\;
}

\caption{\label{alg:feas}Given an initial proof state ($O_{in}$) and
  an empty queue of tactics ($\alpha$), FEAS aims to find a sequence
  of tactics that transforms ($O_{in}$) into the goal state
  (QED). Each proof state is either a set of obligations
  (goal-hypothesis pairs) or an error state. The agent utilizes a
  stack ($st$) to manage proof states, a failure dictionary (Bad($O$))
  to track unproductive tactics, and functions {\tt PROMPTIFY} and
  {\tt PARSETACTIC} to interact with an LLM.  The algorithm proceeds
  by iteratively querying the LLM for tactics, applying them to the
  current proof state, and adjusting its search based on success,
  errors, or lack of progress as determined by a symbolic checker and
  the transition function ($T(O, a)$).}
\end{algorithm}

The FEAS (Functional Equation Automated Solver) agent
(Algorithm~\ref{alg:feas}) builds upon the foundation of the COPRA
framework \cite{ref10}, specializing in the domain of functional
equations.

\subsubsection{Prompt Engineering.} FEAS introduces a key refinement
in the system prompt structure. Rather than directly soliciting a Lean
proof step, FEAS guides a LLM through a multi-stage response
generation process. It prompts the LLM to first articulate a
high-level proof strategy in natural language, then formalize and
translate this strategy into a Lean-compatible proof.

\begin{figure}[ht]
  \centering
  \includegraphics[width=\linewidth]{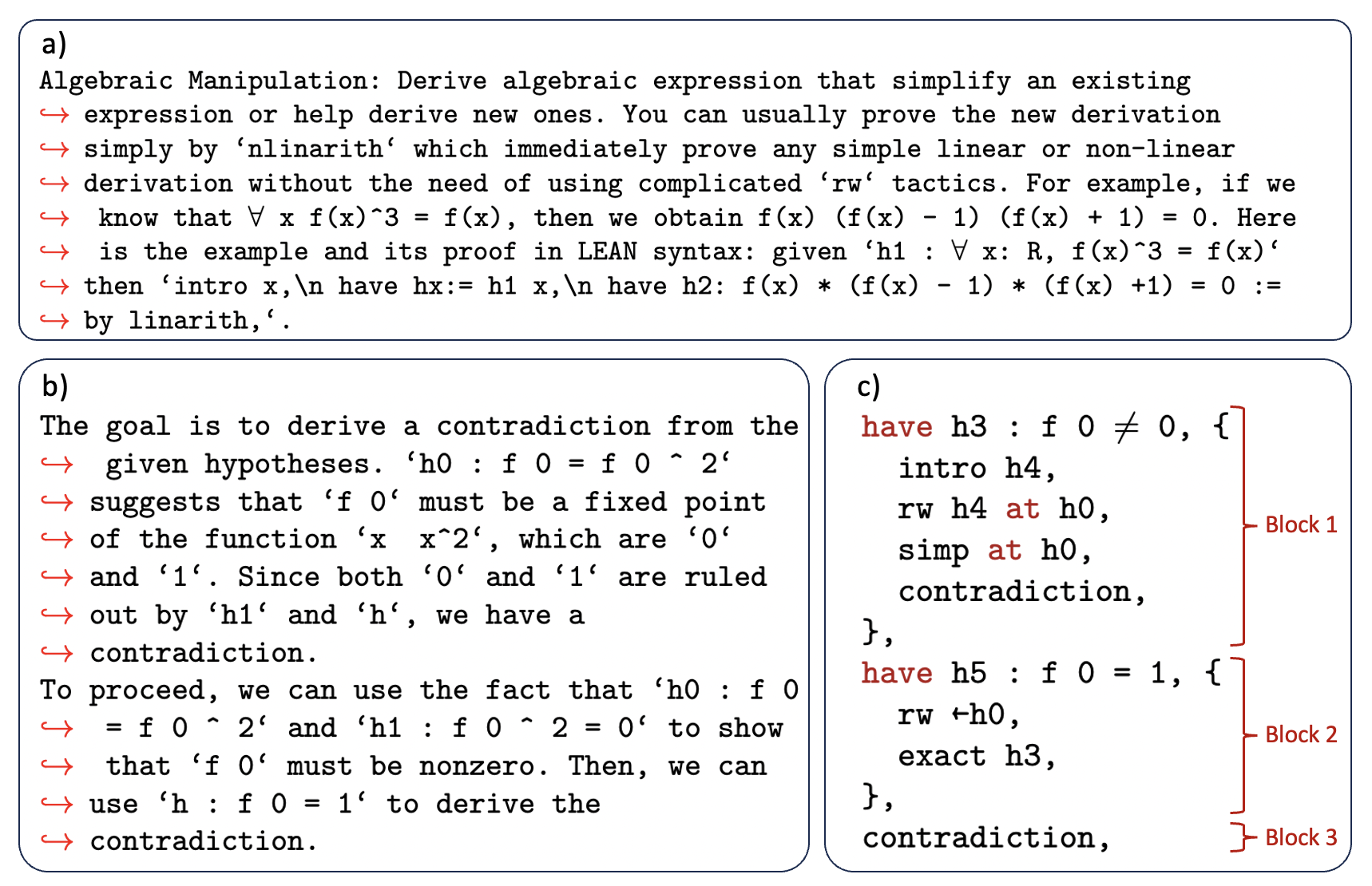}
  \caption{Example of FEAS's Prompting and Response Generation for a
    Functional Equation Problem.  (a) An example of domain-specific
    heuristic included in the system prompt. (b) Natural language
    proof steps generated by the LLM in response to the current proof
    state. (c) The corresponding Lean proof generated by the LLM,
    segmented into blocks for error handling and parsing.}
\end{figure}

\subsubsection{Multi-Block Parsing and Error Handling.} FEAS adopts a
dynamic block-based parsing strategy to manage the multi-line Lean
proofs generated by a LLM. This strategy enhances robustness by
dividing the generated proof into logical blocks based on the
underlying structure of Lean proofs. By processing each block
independently, FEAS can effectively isolate and recover from errors in
specific parts of the proof, potentially salvaging and utilizing valid
proof segments even if the overall proof generated by the LLM is not
entirely correct.

\subsubsection{Automatic Tactic Application.} After either successful
parsing of all blocks or encountering an error, FEAS attempts the
automatic application of the {\tt nlinarith} tactic --- which can
simplify proofs by automatically handling complex algebraic
manipulations that would otherwise need to be done manually. Successful
application incorporates this step into the proof, otherwise, it is
omitted. This provides automation and taps into the power of Lean's
built-in tactics.

\subsubsection{Domain-Specific Heuristics.} FEAS integrates functional
equation heuristics \cite{ref14} directly into the system prompt
alongside Lean syntax examples. These heuristics encompass
substitution-based simplification, techniques for proving bijectivity,
exploitation of symmetry and involution, and the utilization of
induction over natural and rational numbers. This integration aims to
guide the LLM towards generating more relevant and successful proof
strategies within this specific problem domain.

\begin{table}[t]
\centering
\renewcommand{\arraystretch}{1.5} 
\begin{tabularx}{\textwidth}{Xcccc}
\toprule
Algorithm & Few Shots & COPRA & FEAS & FEAS+Heuristics \\
\midrule
GPT & 50.0\% (50.0\%) & 77.78\% (77.78\%) & 80.56\% (83.33\%) & \textbf{86.11}\% (\textbf{94.44}\%) \\
Gemini & 33.34\% (38.89\%) & 52.78\% (61.11\%) & 80.56\% (\textbf{88.89}\%) & \textbf{88.89}\% (\textbf{88.89}\%)\\
Claude & 0\% (0\%) & 83.33\% (83.33\%) & \textbf{91.67}\% (\textbf{100}\%) & 86.11\% (88.89\%) \\
Llama3 & 0\% (0\%) & 50.0\% (50.0\%) & \textbf{75.0}\% (\textbf{77.78}\%) & 63.89\% (66.67\%) \\
\bottomrule
\end{tabularx}
\caption{\label{tab:performance-comparison}Performance comparison of
  pass@1 (pass@2) on the simple tier of FunEq dataset}
\begin{tabularx}{\textwidth}{Xcccc}
\toprule
Algorithm & Few Shots & COPRA & FEAS & FEAS+Heuristics \\
\midrule
GPT & 0\% (0\%) & 0\% (0\%) & 6.67\% (\textbf{13.33}\%) & \textbf{10.0}\% (\textbf{13.33}\%) \\
Gemini & 0\% (0\%) & 6.67\% (6.67\%) & \textbf{10.0}\% (\textbf{13.33}\%) & 3.33\% (6.67\%)\\
Claude & 0\% (0\%) & 6.67\% (6.67\%) & \textbf{13.33}\% (\textbf{13.33}\%) & \textbf{13.33}\% (\textbf{13.33}\%) \\
Llama3 & 0\% (0\%) & 0\% (0\%) & 0\% (0\%) & \textbf{6.67}\% (\textbf{6.67}\%) \\
\bottomrule
\end{tabularx}
\caption{\label{tab:revised-performance-comparison}Performance comparison of
  pass@1 (pass@2) on the intermediate tier of FunEq dataset}
\end{table}

\section{Evaluation}

We conduct a series of experiments to evaluate the performance of the
FEAS agent and the impact of incorporating domain-specific
heuristics. Our evaluation includes comparisons across four different
LLMs: GPT-4 Turbo, Gemini-1.5-Pro, Claude-3.5-Sonnet, and Llama3
70b. We evaluate FEAS agents against two baselines: Few-Shots and
COPRA, the original in-context learning agent, which serves as points
of comparison. We assess FEAS in two distinct configurations: one with
the integrated domain-specific functional equation heuristics and one
without.

The experiments are performed on the simple and intermediate tiers of
the FunEq dataset. To gauge performance on more complex problems, we
further evaluate all agents on the A1 subset of the hard dataset,
which consists of the easiest shortlisted algebra problems from each
corresponding IMO year. To control for potential variability in LLM
responses, we execute each experiment twice on the simple and
intermediate datasets. Due to computational resource constraints, we
limit our evaluation to a single run on the A1 subset.

In all experiments, we impose a maximum limit of 60 LLM queries and a
timeout of 720 seconds. The LLMs are used with a temperature setting
of 0, prioritizing deterministic responses.  Performance is assessed
using Pass@1 and Pass@2 metrics, representing success on the first and
second attempts, respectively. For the simple and intermediate
datasets, Pass@1 is calculated as an average of the results of the two
runs.

\subsection{Results}

Tables \ref{tab:performance-comparison} and
\ref{tab:revised-performance-comparison} show our evaluation across
the Simple and Intermediate tiers of FunEq. All combinations of agents and LLMs fail to generate proofs on the Hard tier of FunEq. On the Simple
dataset, FEAS agents consistently achieves the highest success rates
across all evaluated LLMs. FEAS with integrated heuristics on GPT and
Gemini achieves the highest performance, demonstrating the efficacy of
domain-specific knowledge.  However, Claude and Llama3, without
heuristics, shows superior performance on this dataset, suggesting
that in certain LLM configurations, heuristics may misguide proof
search.

On the more challenging Intermediate dataset, the challenge increases
substantially, with all approaches showing lower success
rates. However, FEAS agents again consistently ranks highest in
performance, highlighting its ability to navigate more complex
functional equation proofs. Again, in some cases, Gemini, FEAS
performs better without heuristics. Furthermore, all methods fail to generate proofs on the Hard tier of FunEq, indicating that significant challenges remain in automated theorem proving for functional
equations.

\section{Conclusion}

Our experiments establish the FEAS agent as an advancement in
automated theorem proving for functional equations. FEAS refines
prompting, parsing, and integration of domain-specific heuristics
demonstrate improvement over baselines. While results on the Simple
dataset are encouraging, performance on the Intermediate and Hard
datasets highlights the ongoing challenges in this complex domain.

Specifically, the challenges revealed by our evaluation can be
decomposed into two distinct sub-problems: (1) proposing
mathematically useful proof steps, and (2) accurately translating
these high-level steps into the formal language of the theorem prover.
Each of these sub-problems poses its own complexities, requiring
distinct approaches for further improvement.

Several avenues present themselves for future research. The
development of agents tailored to specific sub-tasks within the
framework. Incorporating a broader repertoire of high-level proof
tactics within the LLM's prompting could improve the performance of
generating Lean proof steps. Investigating search algorithms beyond
the currently employed depth-first search has the potential to improve
efficiency and solution discovery. Finally, designing efficient
self-learning mechanisms for FEAS would enable it to continuously
refine its strategies based on both successful and unsuccessful proof
attempts.

\bibliographystyle{splncs04}
\bibliography{paper}
\newpage
\section{Appendix}
\subsection{Qualitative Analysis}

\begin{lstlisting}[caption={FEAS Proof}]
theorem intermediate_funeq_2
(f : $\mathbb{R} \to \mathbb{R}$)
(h_0 : $\forall$ x, f(x + 1) = f(x) + 1)
(h_1 : $\forall$ x, x $\neq$ 0 $\to$ f(1/x) = f(x)/x^2) :
$\forall$ x, x $\neq$ 0 $\to$ f(1 + 1/x) = 1 + f(x)/x^2 :=
begin
    intro x,
    intro hx,
    have h_2 : f (1 / x + 1) = f (1 / x) + 1 := h_0 (1 / x),
    have h_3 : f (1 / x) = f x / x ^ 2 := h_1 x hx,
    rw h_3 at h_2,
    rw add_comm at h_2,
    nlinarith,
end
\end{lstlisting}


\begin{lstlisting}[caption={COPRA Incomplete Proof}]
theorem intermediate_funeq_2
(f : $\mathbb{R} \to \mathbb{R}$)
(h_0 : $\forall$ x, f(x + 1) = f(x) + 1)
(h_1 : $\forall$ x, x $\neq$ 0 $\to$ f(1/x) = f(x)/x^2) :
$\forall$ x, x $\neq$ 0 $\to$ f(1 + 1/x) = 1 + f(x)/x^2 :=
begin
    intro x, intro hx,
    rw [h_0 (1 / x), h_1 x hx] at *,
    field_simp [hx],
    rw mul_comm at *,
    rw [$\gets$h_0 (1 / x), h_1 x hx] at *,
    rw [$\gets$h_0 (1 / x), h_1 x hx] at *,
    ring_nf,
end
\end{lstlisting}

To illustrate the distinct strengths of FEAS, we examine a specific
functional equation problem (intermediate\_funeq\_2) where it succeeds
while COPRA does not.  FEAS' solution demonstrates its ability to
generate high-level intermediate proof steps using the {\tt have} tactic,
mirroring a human's approach. This contrasts with COPRA, which focuses
solely on lower-level Lean tactics. FEAS' strategy, guided by the
system prompt instruction to first generate a natural language proof
sketch, leads to a more human-readable and strategically structured
proof.

Furthermore, FEAS' block-by-block parsing successfully handles errors
within individual proof blocks. While all three lines generated by the
LLM contained incorrect tactic applications, FEAS was able to isolate
and utilize the correct proof concepts within each block. This error
recovery mechanism showcases the robustness of FEAS' parsing
strategy.

Interestingly, despite the LLM failing to suggest the final correct
tactic (erroneously proposing {\tt exact h\_2}), FEAS' automated
application of the {\tt nlinarith} tactic successfully concludes the
proof. This demonstrates the complementary nature of FEAS' high-level
reasoning and the underlying theorem prover's automated capabilities.

\end{document}